\documentclass[letterpaper,10pt,conference]{ieeeconf}

\IEEEoverridecommandlockouts
\usepackage{graphicx}
\usepackage{amsmath}
\usepackage{amssymb}
\usepackage{booktabs}
\usepackage{url}
\usepackage{latexml}
\iflatexml
\else
\usepackage{cuted}
\usepackage{capt-of}
\fi

\usepackage{xcolor}
\usepackage[hidelinks]{hyperref}

\definecolor{TacYellow}{HTML}{F4D84A}
\definecolor{TacLightGreen}{HTML}{9CCC52}
\definecolor{TacDarkGreen}{HTML}{28634F}

\newcommand{\method}{TacGooseBumps}
\newcommand{\shortmethod}{TacGB}

\DeclareRobustCommand{\methodmark}{%
  \mbox{%
    \raisebox{-0.14\height}{%
      \includegraphics[height=1.00em]{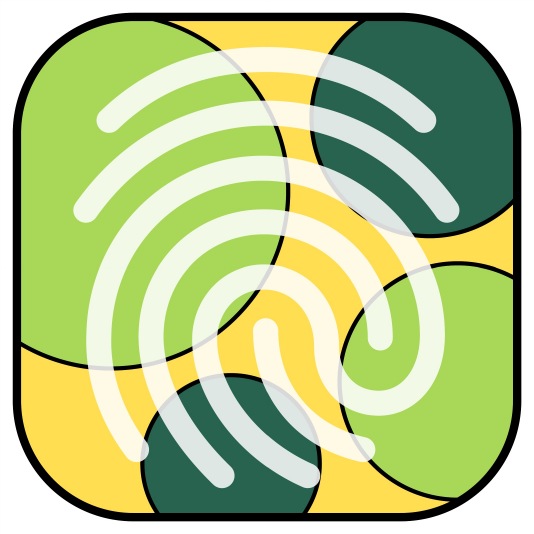}%
    }%
    \hspace{0.28em}%
    \textcolor{TacYellow}{Tac}%
    \textcolor{TacLightGreen}{G}oose%
    \textcolor{TacDarkGreen}{B}umps%
  }%
}

\newcommand{\paperfigure}[2][]{%
  \includegraphics[#1]{#2}}

\title{%
\LARGE\bfseries
\methodmark:\ Retrofitting Normal-Only Tactile Sensors
with Shear Encoding for Learning Contact-Rich Manipulation
}

\iflatexml
\renewcommand{\methodmark}{TacGooseBumps}
\author{%
\IEEEauthorblockN{Wenjie Li}
\IEEEauthorblockA{University of California, Berkeley}
\and
\IEEEauthorblockN{Binyu Yang}
\IEEEauthorblockA{University of California, Berkeley}
\and
\IEEEauthorblockN{Yuxin Chen}
\IEEEauthorblockA{University of California, Berkeley}
\and
\IEEEauthorblockN{Ambrose Wang}
\IEEEauthorblockA{Saratoga High School}
\and
\IEEEauthorblockN{Masayoshi Tomizuka}
\IEEEauthorblockA{University of California, Berkeley}
}
\else
\author{
Wenjie Li$^{*,1}$,
Binyu Yang$^{*,1}$,
Yuxin Chen$^{1}$,
Ambrose Wang$^{2}$,
and Masayoshi Tomizuka$^{1,\#}$%
\thanks{$^{*}$Equal contribution.}%
\thanks{$^{1}$University of California, Berkeley.}%
\thanks{$^{2}$Saratoga High School.}%
\thanks{$^{\#}$Corresponding author:
{\ttfamily\small tomizuka@berkeley.edu}.}%
}

\fi

\begin{document}
\maketitle
\iflatexml
\noindent Wenjie Li and Binyu Yang contributed equally.
Corresponding author: Masayoshi Tomizuka (\texttt{tomizuka@berkeley.edu}).
\par
\fi
\thispagestyle{empty}
\pagestyle{empty}

\iflatexml
\begin{figure*}[t]
    \centering
    \paperfigure[width=\textwidth]{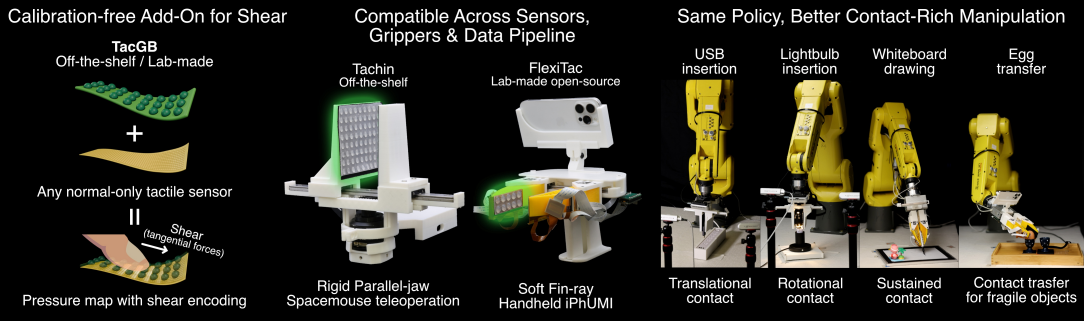}
    \caption{\textbf{An electronics-free passive surface turns normal pressure maps into shear-sensitive observations for policy learning.}
    \shortmethod{} is a thin array of elastomeric domes that can adhere to any existing normal-only tactile sensor without modifying its electronics.
    Under tangential shear load, each dome tilts and redistributes pressure across neighboring taxels, producing direction-dependent spatial patterns that an end-to-end policy can use without force calibration or explicit shear reconstruction.
    We instantiate the same principle with off-the-shelf cabinet bumpers and molded domes, integrate it with normal-only tactile sensors, and evaluate the resulting observations in contact-rich imitation learning across two data-collection pipelines and four contact-rich manipulation tasks.
    }
    \label{fig:overview}
\end{figure*}
\else
\begin{strip}
    \centering
    \vspace{-19mm}
    \paperfigure[width=\textwidth]{figures/Fig_1.pdf}
    \captionof{figure}{\textbf{An electronics-free passive surface turns normal pressure maps into shear-sensitive observations for policy learning.}
    \shortmethod{} is a thin array of elastomeric domes that can adhere to any existing normal-only tactile sensor without modifying its electronics.
    Under tangential shear load, each dome tilts and redistributes pressure across neighboring taxels, producing direction-dependent spatial patterns that an end-to-end policy can use without force calibration or explicit shear reconstruction.
    We instantiate the same principle with off-the-shelf cabinet bumpers and molded domes, integrate it with normal-only tactile sensors, and evaluate the resulting observations in contact-rich imitation learning across two data-collection pipelines and four contact-rich manipulation tasks.
    }
    \label{fig:overview}
\end{strip}
\fi

\begin{abstract}

Contact-rich policies often fail because distinct physical states look alike yet require different actions.
Cameras may not reveal whether a connector is aligned or fully seated, while many normal-only tactile sensors can miss the tangential interactions perpendicular to the grasping direction that distinguish these states.
We ask whether a learning policy needs calibrated shear measurements, or only a repeatable observation that separates shear-dependent contact states.
We introduce \method{} (\shortmethod), a passive domed film that mechanically encodes tangential loading as pattern changes in an existing sensor's pressure map.
Tangential loading tilts each dome and redistributes pressure across its footprint; an end-to-end policy consumes the resulting maps without added electronics, force reconstruction, or taxel-level dome alignment.
Across four imitation-learning tasks and two data-collection pipelines, \shortmethod{} improves goal attainment, efficiency, and contact quality: insertion success increases by up to 36 percentage points, and successful insertions are completed faster, while fragile-object placement becomes gentler and drawing becomes more continuous and straight.
Signal, stage-wise, failure-mode, and trajectory analyses link these gains to contact regimes in which task-relevant tangential interactions are poorly resolved by vision and normal pressure alone.
Together, these results show that shear need not be measured metrically to benefit robot learning; it can instead be mechanically encoded without changing the underlying tactile sensor or the policy's pressure-map input format.
Project website: \href{https://jeffwli.github.io/tacgb/}{https://jeffwli.github.io/tacgb/}.

\end{abstract}

\section{Introduction}

Contact-rich manipulation is not merely about generating a trajectory through visible poses; it requires distinction of a sequence of contact events.
During USB insertion, for example, the wrist image changes little as the connector moves from touching the faceplate, to entering the socket, to reaching the mechanical stop.
The correct action nevertheless changes at each event: search laterally, advance, then stop and release.
Tangential contact forces change immediately across these transitions, whereas vision and grasp-normal pressure can remain nearly unchanged.
A policy that cannot observe this change faces contact-state aliasing: distinct physical states appear similar yet require different actions~\cite{lee2019making}.

Touch can resolve this ambiguity, but common tactile sensing hardware presents a practical trade-off.
Vision-based tactile sensors provide rich deformation fields, including shear, but their optical path adds hardware volume and image-processing overhead \cite{yuan2017gelsight, lambeta2020digit}.
Flexible electronic skins (e-skins) are thin, conformal, and high-frequency, making them attractive for existing grippers and portable data-collection devices \cite{zhu2026touch, wang2023neuromorphic}.
Yet many accessible pressure arrays primarily report normal pressure and provide little direct information about tangential force or torsion---precisely the interactions that signal slip, support, alignment, and rotational constraint.

Existing multidirectional e-skins often solve this limitation through tightly co-designed three-dimensional mechanical structures and electronic sensing elements, followed by calibration against a standardized force/torque sensor~\cite{liu2024three, boutry2018hierarchically}. 
That route is appropriate when the output must be a metric force vector.
Robot learning poses a different question: does a policy need calibrated shear, or only a repeatable observation that makes shear-dependent contact states distinguishable?

\begin{figure}[t]
    \centering
    \paperfigure[width=0.95\linewidth]{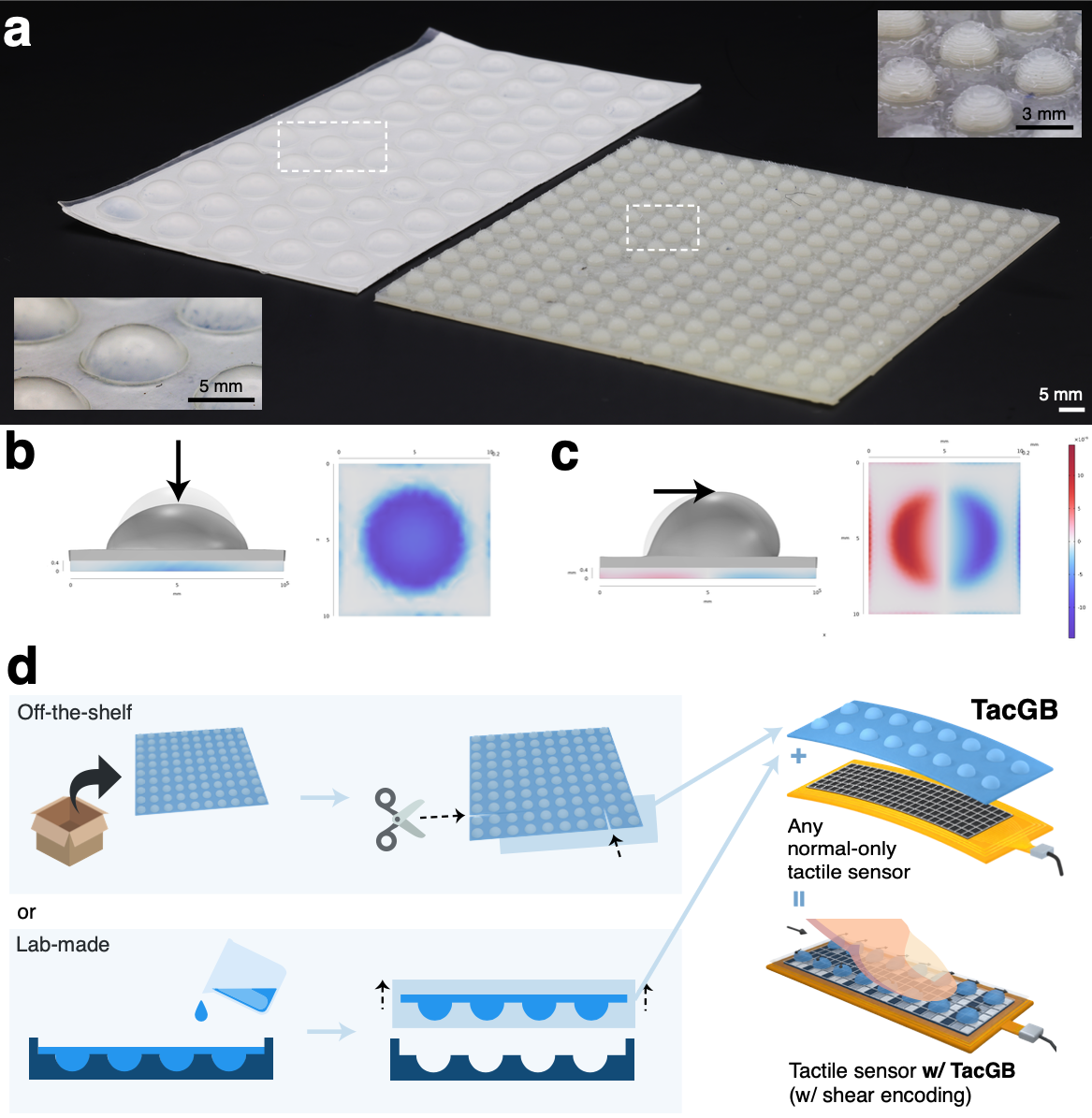}
    \caption{\textbf{Design and integration of TacGB.} (a) Off-the-shelf and lab-made TacGB; insets show magnified dome structures. (b,c) Finite element analysis (\textit{COMSOL}) of a single dome on a normal-only tactile sensor under normal (b) and shear (c) loading. The colorbar shows normal strain in the sensor; dome deformation is magnified by $6000\times$ for better visibility. Left: side view of the whole \shortmethod{}-sensor stack. Right: bottom view of the sensor. (d) Integration procedure. Lab-made TacGB is molded to a target size and shape; off-the-shelf TacGB is cut to fit. Either interface adheres directly to the surface of an existing normal-only tactile sensor. }
    \label{fig:mechanism}
\end{figure}

\textbf{\method{} is NOT a new sensor; it is a passive mechanical observation interface that adds on to existing normal-only tactile sensors without changing their electronics.}
The interface consists of a thin array of elastomeric domes (Fig.~\ref{fig:mechanism}a) that tilt under tangential loading, concentrating normal pressure at the leading edge of each footprint and unloading its trailing edge (Fig.~\ref{fig:mechanism}c).
The underlying sensor therefore records a pressure map whose spatial pattern and temporal evolution depend on shear.
The policy learns directly from these pressure maps, so \shortmethod{} requires neither ground-truth force calibration nor one-to-one dome--taxel alignment.
The interface can be molded for a target geometry or cut from inexpensive, off-the-shelf sheets of cabinet bumpers.

We evaluate this idea as a robot-learning intervention by holding the policy family fixed, changing only the mechanical tactile interface, and structuring the experiments as an increasingly demanding sequence of observability tests.
First, finite-element analysis and a static-load visualization examine whether the interface produces the predicted spatial pressure redistribution.
Second, USB and bayonet lightbulb insertion test whether this redistribution helps policies recognize discrete contact transitions during translational and rotational motion.
Third, egg transfer and whiteboard drawing test whether the same principle improves execution quality under portable demonstrations~\cite{chi2024universal,patel2026behavior}, even when binary success does not capture the difference.
Across all four tasks, the gains concentrate where vision and the bare sensor's normal-pressure readout weakly distinguish physical states, but tangential interaction changes the correct action.

This work makes three contributions:
\begin{itemize}
    \item We introduce a passive, electronics-free add-on that encodes tangential interaction in the pressure maps of existing normal-only tactile sensors, without metric force calibration or explicit shear decoding.
    \item We show in controlled comparisons that both off-the-shelf and molded \shortmethod{}s improve end-to-end imitation policies while preserving the policy architecture and pressure-map input modality. We evaluate the approach across commercial and open-source sensors, rigid and compliant grippers, and SpaceMouse-based robot teleoperation and portable hand-held demonstrations.
    \item We connect improvements in both task success and contact quality to contact-state observability through signal dynamics, substage survival, failure modes, external load traces, and executed trajectories.
\end{itemize}

\section{Related Work}

\subsection{Tactile sensing for manipulation}

Tactile sensors span vision-based tactile sensors~\cite{zhao2025polytouch, lambeta2020digit}, magnetic~\cite{yan2021soft, bhirangi2025anyskin} and capacitive skins~\cite{ha_soft_2022, wistreich2025dexskin}, piezoresistive arrays~\cite{ma2017highly, yun2026multiscale}, and other transduction mechanisms~\cite{oh2020scalable, qiu2024quantitative}.
Vision-based tactile sensors offer high spatial resolution and expose shear through marker or surface motion, but their cameras and illumination add bulk and make integration into existing hardware more invasive~\cite{yuan2017gelsight}. Magnetic tactile sensors can likewise recover multidirectional deformation, yet the Hall-sensor array and its carrier PCB beneath the compliant magnetic layer also occupy volume and form a rigid backing~\cite{yan2021soft, bhirangi2025anyskin}.
Thin electronic arrays remain attractive for existing grippers and portable data-collection tools because of their conformability, yet most accessible variants are sensitive primarily to normal pressure~\cite{wang2023neuromorphic}.
\shortmethod{} does not replace these sensor families; it provides a retrofit for existing normal-only tactile sensors that expands what their native readouts can encode.

\subsection{Mechanical transduction of shear}

Structured skins can convert tangential loading into differential normal response using domes \cite{liu2024three}, pyramids \cite{boutry2018hierarchically, yun2026multiscale}, or other compliant features \cite{oh2020scalable, lee2011real, gloumakov2024fast}.
Prior systems commonly fabricate these features as part of a sensor-specific stack and calibrate their outputs against standardized force/torque sensors to recover force components.
Our objective is different: we separate the mechanical encoding layer from the sensing electronics and optimize its output for downstream policy learning rather than metric force reconstruction.
In our tested integrations, the removable layer is attached without registering individual domes to taxels, allowing the sensing electronics to remain unchanged.

\subsection{Visuotactile policy learning}

Recent robot-learning systems demonstrate that tactile observations improve manipulation under occlusion, fragile-object handling, and precise contact \cite{niu2026t, xue2025reactive, zhu2026touch, yin2023rotating, kerr2022self}. 
Most compare a tactile sensor against a vision-only baseline or propose a new multimodal representation \cite{heng2025vitacformer, park2026tactx, lee2019making}.
We instead study a controlled change to the tactile observation itself: can a passive surface make the pressure-map input from the same normal-only tactile sensor more useful to an otherwise unchanged end-to-end policy?
This framing places \shortmethod{} between sensor design and representation learning: mechanical structures create the informative pattern, while the policy decides how to use it.

\section{Mechanically Encoding Shear for Learning}
\label{sec:method}

\subsection{Design objective}

The goal is not to reconstruct a calibrated three-axis force vector, but to make task-relevant tangential loading observable in the native pressure-map stream of existing normal-only tactile sensors. \shortmethod{} achieves this with a dome array that converts such loading into local pressure redistribution on the underlying sensor.

Each dome spans multiple taxels of the underlying pressure array (Fig.~\ref{fig:mechanism}a).
A tangential shear load $F_{\parallel}$ applied above the dome base produces a moment that tilts the dome.
Relative to the unloaded state, pressure increases on the leading side of the footprint and decreases on the trailing side (Fig.~\ref{fig:mechanism}c).
In contrast, a normal load predominantly changes the magnitude of the footprint without the same directional redistribution (Fig.~\ref{fig:mechanism}b).

The policy receives tactile information still in the form of a pressure map $T_t\in\mathbb{R}^{H\times W}$.
We do not estimate $F_{\parallel}$, pair individual taxels with domes, or calibrate the map against a reference load cell.
With a history of pressure maps, the encoder can learn both the spatial differential and its evolution through a contact transition.

\begin{figure}[t]
    \centering
    \paperfigure[width=0.95\linewidth]{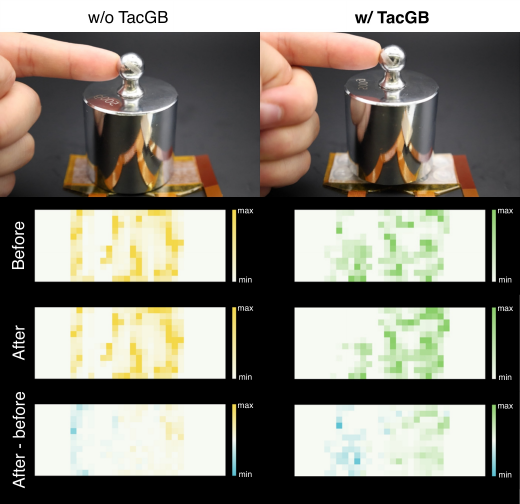}
    \caption{\textbf{Qualitative static-load visualization on \textit{FlexiTac} \cite{huang2026flexitac, huang20243d}.}
 A rightward tangential load is applied to a stationary 500-g mass. Before and after maps and their difference show stronger directional redistribution with \shortmethod{}.}
    \label{fig:visual}
\end{figure}

A qualitative static-load experiment visualizes the same encoding in real sensor data (Fig.~\ref{fig:visual}).
A rightward tangential load is applied to the top of a stationary 500-g mass resting on \textit{FlexiTac} \cite{huang2026flexitac, huang20243d}; static friction prevents gross sliding.
The bare sensor changes little, whereas \shortmethod{} produces a clear signed spatial redistribution.

\subsection{Off-the-shelf and molded interfaces}

\begin{figure}[t]
    \centering
    \paperfigure[width=0.95\linewidth]{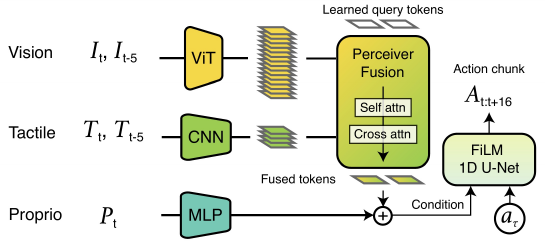}
    \caption{\textbf{Policy architecture.} Fused visual and tactile features, together with proprioception,
condition a flow-matching policy that predicts action chunks.}
    \label{fig:policy}
\end{figure}

The off-the-shelf realization is cut from a sheet of commercially available dome-shaped cabinet bumpers.
Its domes are 6\,mm in diameter with 9\,mm center-to-center spacing.
The molded realization uses VytaFlex~60 silicone~(\textit{Smooth-On}); its 3\,mm domes are arranged at 5\,mm center-to-center spacing so that each dome covers approximately a $2\times2$ group of taxels on the $32\times32$ commercial capacitive sensor~(\textit{Tachin}) used below.
Both variants attach directly over the active sensing area to the \textit{Tachin} sensor.
For the open-source piezoresistive arrays~(\textit{FlexiTac}) integrated with iPhUMI \cite{patel2026behavior, choi2026wild}, we assemble a $12\times32$-taxel sensor and cover it with a $2\times5$ array cut from a different off-the-shelf bumper roll; the domes are 10\,mm in diameter with 12.5\,mm center-to-center spacing.

\begin{figure*}[t]
    \centering
    \paperfigure[width=\textwidth]{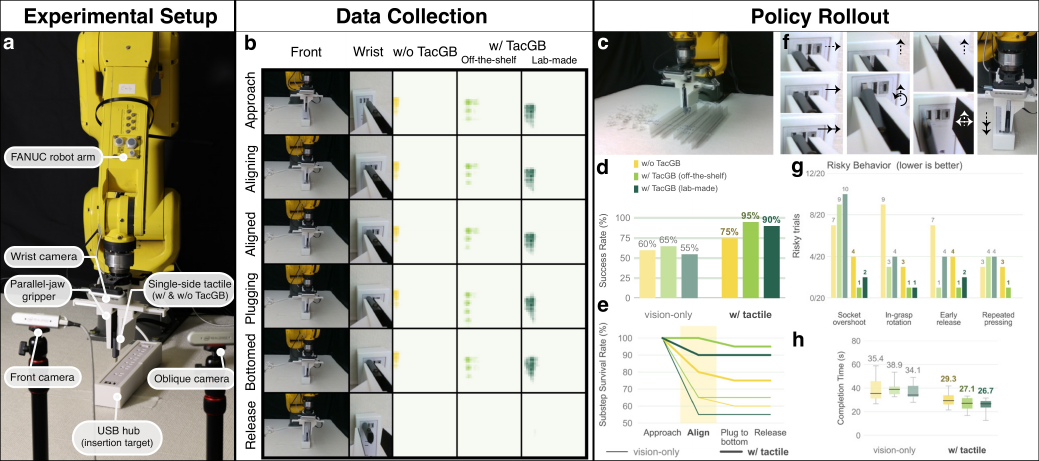}
    \caption{\textbf{USB insertion.} 
    (a) Experimental setup. 
    (b) Representative demonstration stages with front and wrist images and pressure maps from the bare, off-the-shelf, and molded interfaces. 
    (c) Initial states for all evaluation episodes overlaid together. All policies start with the same set of initial states, matched manually with reference images. 
    (d) Success rates over 20 deployment rollouts. 
    (e) Substep survival rates. 
    (f) Typical risky behaviors. From left to right: socket overshoot, in-grasp rotation, early release, and repeated pressing. Dashed arrows indicate desired actions, while solid arrows indicate the observed risky motion.
    (g) Risky behavior counts over 20 deployment rollouts. 
    (h) Task completion-time distributions over successful rollouts.}
    \label{fig:usb}
\end{figure*}
\section{Evaluation I: Recognizing Hidden Contact Transitions}
\label{sec:evaluation}

The rigid-gripper experiments address two questions.
\textbf{Q1:} Does \shortmethod{} improve task success and efficiency for end-to-end policies?
\textbf{Q2:} Do the improvements localize to stages and failure modes that require tangential contact information?

\subsection{Hardware setup, policy, and evaluation protocol}

We mount the $32\times32$ normal-only \textit{Tachin} sensor over an $80\,\mathrm{mm}\times75\,\mathrm{mm}$ area on one finger of a rigid parallel-jaw gripper.
A FANUC LR Mate 200\textit{i}D/7L robot arm is teleoperated with a SpaceMouse for data collection.
The observation contains three RGB views~(front, oblique, and wrist), the tactile pressure map, and 8-D proprioception comprising end-effector pose and gripper openness.
RGB and tactile observations use the current frame and the frame five control steps earlier; proprioception uses the current state~(Fig.~\ref{fig:policy}).

All RGB views share a frozen DINOv2-S encoder~\cite{oquab2023dinov2}.
A compact convolutional network maps the tactile pair to a spatial token grid.
Learned queries aggregate visual and tactile tokens through a Perceiver-style module \cite{jaegle2021perceiver}; the fused representation and an MLP encoding of proprioception condition a FiLM-modulated 1-D U-Net flow-matching head \cite{perez2018film, lipman2022flow}.
The head predicts 16 actions, each containing Cartesian position and Euler-angle deltas plus a binary gripper command, and executes the first eight at inference.

For each task, all policy variants use the same architecture and training schedule.
Initial test configurations are shared across variants and are disjoint from those used for data collection.
We report task success, substage survival, and behaviorally defined failure modes; completion time is reported only for successful trials.
\begin{figure*}[t]
\centering
\paperfigure[width=\textwidth]{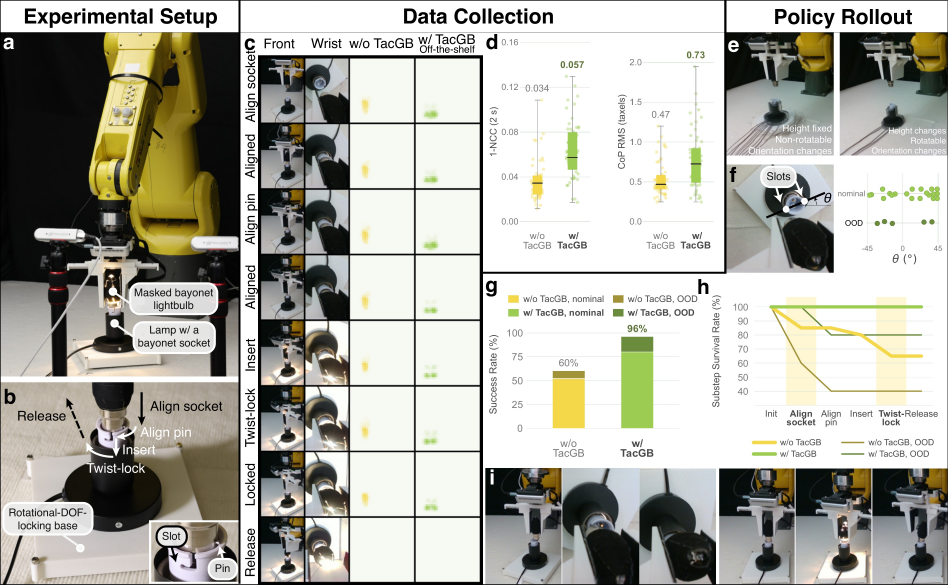}

\caption{\textbf{Lightbulb insertion.}
(a) Experimental setup. The LED lightbulb is masked to protect the teleoperator from glare.
(b) Specification of the substeps and detailed mechanical constraints. Inset: Zoomed view of the bayonet structure on the lightbulb and the lamp socket.
(c) Representative observations from collected demonstrations during alignment, insertion, twist-locking, and release.
(d) Tactile dynamics statistics of collected demonstrations. Left: pressure-map dissimilarity $1-\mathrm{NCC}$ over $2\,\mathrm{s}$. Right: planar center-of-pressure RMS; printed values are medians.
(e) Twenty nominal initializations with fixed socket height and locked rotation, plus five OOD initializations with changed height and free socket rotation.
(f) Distribution of initial placement angle $\theta$.
(g) Success rates over all 25 rollouts.
(h) Substep survival rates. Substeps that cause major drops in survival rates are highlighted.
(i) Typical failures. Left: off-center alignment followed by irreversible in-grasp rotation. Right: release after vertical insertion without twist locking. The lightbulb will pop up due to the spring recovery beneath the electrodes.}
\label{fig:bulb}
\end{figure*}
\subsection{USB insertion: contact transitions in translation}
\label{sec:usb}

The robot must approach a power strip, align a pre-grasped USB connector with a socket, insert it to the mechanical stop, release, and retract.
This task is deliberately visually aliased (Fig. \ref{fig:usb}a,b): faceplate contact, partial alignment, and full seating can look similar from the cameras but demand different actions.
We collect 100 demonstrations for each of the three tactile interfaces: w/o \shortmethod{} (the unmodified bare normal-only tactile sensor), w/ \shortmethod{} (off-the-shelf), and w/ \shortmethod{} (lab-made).
Each learned policy is evaluated on the same 20 initial configurations~(Fig. \ref{fig:usb}c).

Fig.~\ref{fig:usb}d shows the task-level result.
With tactile observations, the bare-array policy succeeds in 15/20 trials (75\%), compared with 19/20 (95\%) for the off-the-shelf interface and 18/20 (90\%) for the lab-made interface.
The two geometrically different \shortmethod{} variants perform similarly, suggesting that the benefit does not depend on precise one-to-one dome--taxel alignment.

To test whether the gain is merely a mechanical effect of the \shortmethod{} surface, such as increased friction, we train vision-only policies under all three physical configurations.
Without pressure maps provided as policy input, vision-only policies achieve 60\%, 65\%, and 55\% success for w/o \shortmethod{}, w/ \shortmethod{}~(off-the-shelf), and w/ \shortmethod{}~(lab-made), respectively, compared with 75\%, 95\%, and 90\% when the corresponding pressure maps are available~(Fig.~\ref{fig:usb}d).
Changing the surface mechanics alone therefore produces no consistent improvement and even lowers success in the lab-made condition; the gains emerge only when the policy receives the \shortmethod{}-encoded pressure maps.

To localize where these gains arise, we compare survival rates at each substep. 
\shortmethod{} delivers its largest stage-wise gain at alignment, the principal survival bottleneck for the bare-sensor policy~(Fig.~\ref{fig:usb}e).
This stage is visually ambiguous: the camera view changes little between faceplate contact and socket entry, but the correct action switches from lateral search to advance. Before contact, the gripper supports the connector's weight; faceplate contact partially unloads the gripper; entry into the socket restores the load. Later, bottoming out produces an opposing reaction that indicates when to stop.
Rather than estimating these load transfers explicitly, \shortmethod{} encodes them as pressure-map changes that the policy can distinguish.

With access to \shortmethod{}-encoded pressure maps, the policies show fewer instances of all four risky behaviors in Fig.~\ref{fig:usb}f--g, with two mainly affecting success and two mainly affecting efficiency. In-grasp rotation and premature release lead to failure: repeated faceplate interactions can rotate the USB drive within the grasp, while premature release occurs when the gripper lets go near the socket-aligned position before insertion. Overshoot and continued pressing instead waste time by repeatedly passing the socket-aligned position or pressing after the mechanical stop. Consistent with reductions in these time-wasting behaviors, the mean completion time among successful trials decreases from 29.3\,s w/o \shortmethod{} to 27.1\,s and 26.7\,s w/ \shortmethod{}~(off-the-shelf) and w/ \shortmethod{}~(lab-made), respectively, versus approximately 35\,s for vision-only policies~(Fig.~\ref{fig:usb}h).

\begin{figure*}[t]
    \centering
    \paperfigure[width=\linewidth]{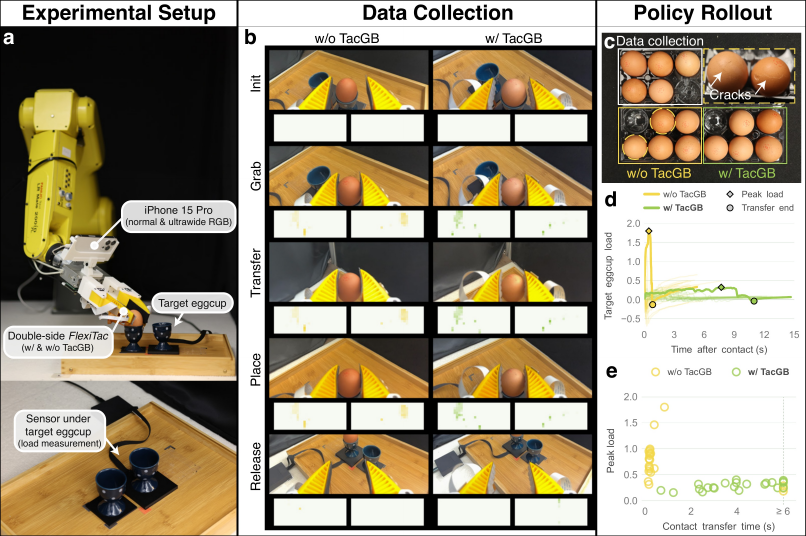}
    \caption{\textbf{Egg transfer.}
    (a) Experimental setup.
    (b) Representative observations from collected demonstrations during grasping, transfer, placement, and release.
    (c) Raw eggs used for data collection and for robot deployment with and without \shortmethod{}; the inset shows visible cracks after deployment without \shortmethod{}.
    (d) Tared target-cup sensor readout after first contact. The peak readout and the post-release minimum used to mark the end of contact transfer are marked out on the representative readouts.
    (e) Peak readout versus contact-transfer duration for all rollouts; sharp peaks and short transfer durations characterize dropping, whereas lower peaks and longer durations characterize controlled placement.}
\label{fig:egg}
\end{figure*}
    
\subsection{Lightbulb insertion: contact transitions in rotation}
\label{sec:bulb}

The second task adds rotational motion.
The robot must bring a pre-grasped bulb to the socket, align two bayonet pins with vertical slots, push against a spring-loaded electrode, rotate the pins into horizontal locking slots, release, and retract (Fig. \ref{fig:bulb}a--c).
Success requires both illumination and mechanical locking; a bulb that is merely inserted without twist lock pops back out when released.

We use the off-the-shelf interface because the USB study found comparable performance for off-the-shelf and lab-made variants.
We collect 55 demonstrations for policy training and evaluate each sensing condition on 20 nominal and five out-of-distribution (OOD) trials.
Nominal trials vary the socket orientation while holding its height and rotational degree of freedom fixed.
In the OOD trials, the socket height changes and its rotational degree of freedom is unlocked (Fig.~\ref{fig:bulb}e--f).

The overall success count increases from 15/25 without \shortmethod{} to 24/25 with it (Fig.~\ref{fig:bulb}g).
On nominal trials, success rises from 13/20 to 20/20.
On the five OOD trials, it rises from 2/5 to 4/5.

\shortmethod{} is designed to make changes in tangential contact more visible in the pressure-map sequence. We test this prediction using pairs of lightbulb-demonstration frames separated by 2\,s (Fig.~\ref{fig:bulb}d). Normalized cross-correlation (NCC) measures the similarity between two pressure maps, so $1-\mathrm{NCC}$ increases as their spatial patterns become more different. With \shortmethod{}, the median and mean values of $1-\mathrm{NCC}$ are $1.68\times$ and $1.70\times$ higher, respectively. The RMS displacement of the center of pressure (CoP) captures the extent to which the pressure distribution shifts; its median and mean are $1.55\times$ and $1.43\times$ higher. Together, both metrics show more distinguishable spatiotemporal pressure-map changes with \shortmethod{}.

\shortmethod{} improves the two stages where the contact direction changes while the visual scene remains similar.
Rim contact adds an upward tangential interaction; pin entry into a vertical slot abruptly removes part of that support; bottoming out increases the interaction from the spring-loaded electrode; and rotating into the horizontal locking slot redirects the tangential interaction before the terminal stop.
Without \shortmethod{}, half of the observed failures arise during socket alignment and another 30\% during twist locking~(Fig.~\ref{fig:bulb}h).
Representative rollouts repeatedly press at an off-center pose until the bulb rotates irreversibly in the grasp, or release after vertical insertion without completing the twist-locking process, which leads to the lightbulb popping up again due to the spring recovery beneath the electrodes~(Fig.~\ref{fig:bulb}i).
Both errors occur when visually similar states require different actions, matching the contact-state aliasing hypothesis.

\section{Evaluation II: From Binary Success to Contact Quality}
\label{sec:umi}

The insertion experiments ask whether shear encoding helps a policy complete a task.
The portable-demonstration experiments ask a harder question: can it improve how the task is completed even when binary success is unchanged?
We study two complementary forms of contact quality---a discrete transfer of support during fragile placement and sustained two-directional interaction during drawing---using human-operated iPhUMI demonstrations.

\subsection{Tactile UMI setup and protocol}

We integrate $12\times32$ \textit{FlexiTac} normal-only tactile arrays into a hand-held \textit{iPhUMI} gripper and its robot-mounted counterpart.
Each sensing surface is evaluated either bare or with a $2\times5$ \shortmethod{} array cut from an off-the-shelf bumper roll; the domes are 10\,mm in diameter with 12.5\,mm center-to-center spacing.
The observation contains two wrist RGB views from the iPhone main and ultrawide cameras, left and right pressure maps, and proprioception obtained from ARKit during demonstration and from the robot during deployment.
RGB and tactile inputs use the current frame and the frame five control steps earlier.
Proprioception uses the current and previous control steps, each represented by a 16-D vector containing current-relative TCP position and rotation, episode-start-relative rotation, and absolute gripper width.
The policy retains the architecture in Section~\ref{sec:evaluation}; its head predicts 16 actions and executes the first eight, with each action comprising Cartesian-position and 6-D-rotation deltas plus continuous gripper width.

For every task and sensing condition, two operators collect 100 demonstrations.
All policies share the architecture, training schedule, and evaluation initializations; test configurations are disjoint from data collection.

\subsection{Fragile placement: observing support transfer before release}

\begin{figure}[!t]
    \centering
    \paperfigure[width=\linewidth]{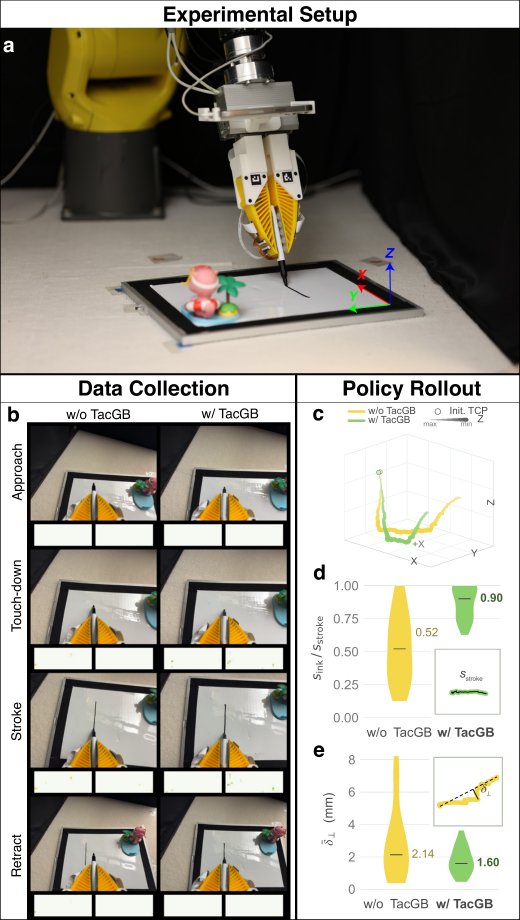}
    \caption{\textbf{Whiteboard drawing.}
    (a) Experimental setup and whiteboard-coordinate definition.
    (b) Representative observations from collected demonstrations during approach, touchdown, stroke, and retraction. The tactile streams are tared during the first 3\,s of each recording or rollout while the marker is already grasped, so the plotted values encode changes from the initial grasp rather than absolute grasp pressure.
    (c) Representative rollout TCP trajectories in whiteboard coordinates.
    (d) Ink coverage ratio $s_{\mathrm{ink}}/s_{\mathrm{stroke}}$ across 20 rollouts per policy. Inset: board-plane projection of the TCP trajectory with \shortmethod{} in (c) between touchdown and retraction, illustrating the definition of $s_{\mathrm{stroke}}$.
    (e) Mean perpendicular trajectory deviation $\bar{\delta}_{\perp}$ across the same rollouts. Inset: board-plane projection of the TCP trajectory without \shortmethod{} in (c) between touchdown and retraction, illustrating the definition of $\delta_{\perp}$.}
    \label{fig:wb}
\end{figure}

The robot grasps a raw egg from a source cup, transfers it to a target egg cup, releases it, and retracts.
Before placement, the gripper supports the egg through tangential contact; after cup contact, that support should transfer gradually to the target cup before release (Fig.~\ref{fig:egg}a--c).
The cameras weakly reveal this transition, and binary success cannot distinguish a controlled placement from a short drop that leaves the egg in the cup.
We therefore place a commercial \textit{Tachin} sensor beneath the target cup and use its tared aggregate readout as an external load proxy, not as an input to the policy.

We collect 100 demonstrations per sensing condition using the same five training eggs.
For evaluation, each policy performs five rollouts on each of five eggs, for 25 rollouts per condition; separate but physically similar egg sets are used across conditions to avoid transferring accumulated damage across rollouts of different policies.
We define contact-transfer duration from the first sustained cup contact to the post-release minimum in the cup-sensor signal, after which the signal relaxes without robot contact~(Fig.~\ref{fig:egg}d).

Both policies achieve 25/25 binary successes, but their contact quality differs sharply.
Without \shortmethod{}, premature unloading of the gripper produces impact-like traces: the target-cup readout rises rapidly to a high peak and reaches the transfer endpoint soon afterward.
Two of the five eggs used in this condition develop visible cracks despite remaining intact in every rollout.
With \shortmethod{}, mean peak readout decreases by 66\%, from 0.77 to 0.26, while mean contact-transfer duration increases from 0.99\,s to 3.84\,s (Fig.~\ref{fig:egg}e).
Thus, \shortmethod{} changes a behavior that binary success misses: waiting for stable support before releasing the fragile object.

\subsection{Whiteboard drawing: sustaining two-directional interaction}
The robot approaches a horizontal whiteboard, brings a marker into contact, draws a human-demonstrated numeral ``1'' approximately along the board's $X$-axis, and retracts (Fig.~\ref{fig:wb}a--b).
Unlike insertion and placement, this task requires continuous regulation rather than recognition of a few discrete events.
In the finger frame, the marker--board reaction has two action-relevant components: a $Z$-direction interaction that accompanies maintained tip contact and an $X$-direction interaction associated with friction during the stroke.
The bare array is dominated by grasp-normal~($Y$-direction) loading, whereas \shortmethod{} makes changes along both tangential directions available in the pressure-map sequence.

We evaluate each policy in 20 rollouts and analyze the TCP segment between touchdown and retraction (Fig.~\ref{fig:wb}c).
To quantify maintained contact, we project the TCP path onto the board plane, measure its arc length $s_{\mathrm{stroke}}$, and measure the portion covered by ink, $s_{\mathrm{ink}}$.
The coverage ratio $s_{\mathrm{ink}}/s_{\mathrm{stroke}}$ increases by 73.1\% with \shortmethod{} (Fig.~\ref{fig:wb}d).
To quantify straightness, we compute the mean perpendicular distance $\bar{\delta}_{\perp}$ from the contact trajectory to the chord joining its endpoints.
This deviation decreases by 25.2\% with \shortmethod{} (Fig.~\ref{fig:wb}e).
The two changes are complementary: the policy maintains marker contact while suppressing lateral wandering, rather than improving one by sacrificing the other.

Together, the UMI experiments extend the insertion result from discrete goal attainment to execution quality.
Across a deformable gripper, portable demonstrations, and two different contact regimes, shear-encoded pressure maps help the policy regulate when to release and how to sustain contact.

\section{Discussion and Limitations}

Our results support a deliberately scoped claim: a mechanical interface can make task-relevant tangential interactions more observable to an end-to-end policy using normal-only tactile sensors.
They do not establish calibrated shear-force sensing.
However, signal magnitude depends on dome geometry, material, sensor resolution, and attachment, so changing an interface may require new policy data even though it does not require force calibration.

Generality is demonstrated across two sensors, two interface geometries, rigid and compliant parallel-jaw grippers, and two data-collection pipelines.
It is not yet established for dexterous hands, other transduction mechanisms, or substantially different taxel pitches.
The UMI system also timestamps phone and tactile streams on separate clocks and aligns them post hoc using a session-level offset estimated from gripper-width--pressure correlation.
A shared acquisition clock would remove uncertainty from clock error, contact delay, and intermittent AR-tag detection.

Finally, the evaluation should also be read at the correct statistical level.
Repeated rollouts quantify deployment behavior but do not replace independently trained policy seeds.
Future studies should separate training, object, and rollout variability while systematically varying dome geometry, material, and layout to better understand the underlying mechanics.

\section{Conclusion}

\method{} upgrades what an existing normal-only tactile sensor can reveal without replacing its electronics or reconstructing a force vector.
Its passive domes convert tangential interaction into direction-dependent pressure-map changes that an unchanged end-to-end policy can learn directly.
Across rigid-gripper insertion and portable-demonstration tasks, these observations improve discrete success where contact states are visually aliased and improve execution quality where success alone hides dropping, intermittent contact, or lateral drift.
The broader result is a design principle for robot learning: a useful sensing interface need not recover every physical quantity; it must make action-relevant distinctions observable and repeatable.

\bibliographystyle{IEEEtran}
\bibliography{references}

\begin{thebibliography}{10}
\providecommand{\url}[1]{#1}
\csname url@samestyle\endcsname
\providecommand{\newblock}{\relax}
\providecommand{\bibinfo}[2]{#2}
\providecommand{\BIBentrySTDinterwordspacing}{\spaceskip=0pt\relax}
\providecommand{\BIBentryALTinterwordstretchfactor}{4}
\providecommand{\BIBentryALTinterwordspacing}{\spaceskip=\fontdimen2\font plus
\BIBentryALTinterwordstretchfactor\fontdimen3\font minus \fontdimen4\font\relax}
\providecommand{\BIBforeignlanguage}[2]{{%
\expandafter\ifx\csname l@#1\endcsname\relax
\typeout{** WARNING: IEEEtran.bst: No hyphenation pattern has been}%
\typeout{** loaded for the language `#1'. Using the pattern for}%
\typeout{** the default language instead.}%
\else
\language=\csname l@#1\endcsname
\fi
#2}}
\providecommand{\BIBdecl}{\relax}
\BIBdecl

\bibitem{lee2019making}
M.~A. Lee, Y.~Zhu, K.~Srinivasan, P.~Shah, S.~Savarese, L.~Fei-Fei, A.~Garg, and J.~Bohg, ``Making sense of vision and touch: Self-supervised learning of multimodal representations for contact-rich tasks,'' in \emph{2019 International conference on robotics and automation (ICRA)}.\hskip 1em plus 0.5em minus 0.4em\relax IEEE, 2019, pp. 8943--8950.

\bibitem{yuan2017gelsight}
W.~Yuan, S.~Dong, and E.~H. Adelson, ``Gelsight: High-resolution robot tactile sensors for estimating geometry and force,'' \emph{Sensors}, vol.~17, no.~12, p. 2762, 2017.

\bibitem{lambeta2020digit}
M.~Lambeta, P.-W. Chou, S.~Tian, B.~Yang, B.~Maloon, V.~R. Most, D.~Stroud, R.~Santos, A.~Byagowi, G.~Kammerer \emph{et~al.}, ``Digit: A novel design for a low-cost compact high-resolution tactile sensor with application to in-hand manipulation,'' \emph{IEEE Robotics and Automation Letters}, vol.~5, no.~3, pp. 3838--3845, 2020.

\bibitem{zhu2026touch}
X.~Zhu, B.~Huang, and Y.~Li, ``Touch in the wild: Learning fine-grained manipulation with a portable visuo-tactile gripper,'' \emph{Advances in Neural Information Processing Systems}, vol.~38, pp. 153\,783--153\,812, 2026.

\bibitem{wang2023neuromorphic}
W.~Wang, Y.~Jiang, D.~Zhong, Z.~Zhang, S.~Choudhury, J.-C. Lai, H.~Gong, S.~Niu, X.~Yan, Y.~Zheng \emph{et~al.}, ``Neuromorphic sensorimotor loop embodied by monolithically integrated, low-voltage, soft e-skin,'' \emph{Science}, vol. 380, no. 6646, pp. 735--742, 2023.

\bibitem{liu2024three}
Z.~Liu, X.~Hu, R.~Bo, Y.~Yang, X.~Cheng, W.~Pang, Q.~Liu, Y.~Wang, S.~Wang, S.~Xu \emph{et~al.}, ``A three-dimensionally architected electronic skin mimicking human mechanosensation,'' \emph{Science}, vol. 384, no. 6699, pp. 987--994, 2024.

\bibitem{boutry2018hierarchically}
C.~M. Boutry, M.~Negre, M.~Jorda, O.~Vardoulis, A.~Chortos, O.~Khatib, and Z.~Bao, ``A hierarchically patterned, bioinspired e-skin able to detect the direction of applied pressure for robotics,'' \emph{Science Robotics}, vol.~3, no.~24, p. eaau6914, 2018.

\bibitem{chi2024universal}
C.~Chi, Z.~Xu, C.~Pan, E.~Cousineau, B.~Burchfiel, S.~Feng, R.~Tedrake, and S.~Song, ``Universal manipulation interface: In-the-wild robot teaching without in-the-wild robots,'' in \emph{Proceedings of Robotics: Science and Systems}, 2024.

\bibitem{patel2026behavior}
A.~Patel, B.~Pekarek, J.~E.~C. Hernandez, and S.~Song, ``Behavior prompting policy: Demonstrations as prompts for manipulation,'' \emph{arXiv preprint arXiv:2606.30457}, 2026.

\bibitem{zhao2025polytouch}
J.~Zhao, N.~Kuppuswamy, S.~Feng, B.~Burchfiel, and E.~Adelson, ``Polytouch: A robust multi-modal tactile sensor for contact-rich manipulation using tactile-diffusion policies,'' in \emph{2025 IEEE International Conference on Robotics and Automation (ICRA)}.\hskip 1em plus 0.5em minus 0.4em\relax IEEE, 2025, pp. 104--110.

\bibitem{yan2021soft}
Y.~Yan, Z.~Hu, Z.~Yang, W.~Yuan, C.~Song, J.~Pan, and Y.~Shen, ``Soft magnetic skin for super-resolution tactile sensing with force self-decoupling,'' \emph{Science Robotics}, vol.~6, no.~51, p. eabc8801, 2021.

\bibitem{bhirangi2025anyskin}
R.~Bhirangi, V.~Pattabiraman, E.~Erciyes, Y.~Cao, T.~Hellebrekers, and L.~Pinto, ``Anyskin: Plug-and-play skin sensing for robotic touch,'' in \emph{2025 IEEE International Conference on Robotics and Automation (ICRA)}.\hskip 1em plus 0.5em minus 0.4em\relax IEEE, 2025, pp. 16\,563--16\,570.

\bibitem{ha_soft_2022}
\BIBentryALTinterwordspacing
K.-H. Ha, H.~Huh, Z.~Li, and N.~Lu, ``Soft {Capacitive} {Pressure} {Sensors}: {Trends}, {Challenges}, and {Perspectives},'' \emph{ACS Nano}, vol.~16, no.~3, pp. 3442--3448, Mar. 2022, publisher: American Chemical Society. [Online]. Available: \url{https://doi.org/10.1021/acsnano.2c00308}
\BIBentrySTDinterwordspacing

\bibitem{wistreich2025dexskin}
S.~Wistreich, B.~Shi, S.~Tian, S.~Clarke, M.~Nath, C.~Xu, Z.~Bao, and J.~Wu, ``{DexSkin}: High-coverage conformable robotic skin for learning contact-rich manipulation,'' in \emph{Proceedings of The 9th Conference on Robot Learning}, ser. Proceedings of Machine Learning Research, vol. 305.\hskip 1em plus 0.5em minus 0.4em\relax PMLR, 2025, pp. 769--793.

\bibitem{ma2017highly}
Y.~Ma, N.~Liu, L.~Li, X.~Hu, Z.~Zou, J.~Wang, S.~Luo, and Y.~Gao, ``A highly flexible and sensitive piezoresistive sensor based on mxene with greatly changed interlayer distances,'' \emph{Nature communications}, vol.~8, no.~1, p. 1207, 2017.

\bibitem{yun2026multiscale}
G.~Yun, Z.~Chen, Z.~Chen, J.~Chen, B.~Zhou, M.~Xiao, M.~Stevens, M.~Chhowalla, and T.~Hasan, ``Multiscale-structured miniaturized 3d force sensors,'' \emph{Nature Materials}, pp. 1--9, 2026.

\bibitem{oh2020scalable}
H.~Oh, G.-C. Yi, M.~Yip, and S.~A. Dayeh, ``Scalable tactile sensor arrays on flexible substrates with high spatiotemporal resolution enabling slip and grip for closed-loop robotics,'' \emph{Science advances}, vol.~6, no.~46, p. eabd7795, 2020.

\bibitem{qiu2024quantitative}
Y.~Qiu, F.~Wang, Z.~Zhang, K.~Shi, Y.~Song, J.~Lu, M.~Xu, M.~Qian, W.~Zhang, J.~Wu \emph{et~al.}, ``Quantitative softness and texture bimodal haptic sensors for robotic clinical feature identification and intelligent picking,'' \emph{Science Advances}, vol.~10, no.~30, p. eadp0348, 2024.

\bibitem{lee2011real}
H.-K. Lee, J.~Chung, S.-I. Chang, and E.~Yoon, ``Real-time measurement of the three-axis contact force distribution using a flexible capacitive polymer tactile sensor,'' \emph{Journal of Micromechanics and Microengineering}, vol.~21, no.~3, p. 035010, 2011.

\bibitem{gloumakov2024fast}
Y.~Gloumakov, T.~M. Huh, and H.~S. Stuart, ``Fast in-hand slip control on unfeatured objects with programmable tactile sensing,'' \emph{IEEE Robotics and Automation Letters}, vol.~9, no.~7, pp. 6059--6066, 2024.

\bibitem{niu2026t}
D.~Niu, Z.~Liu, Z.~Wang, B.~Shao, Z.-H. Yin, A.~Pai, Y.~Sharma, S.~Saravalle, R.~Zheng, J.~Wang \emph{et~al.}, ``{T-Rex}: Tactile-reactive dexterous manipulation,'' in \emph{Conference on Robot Learning (CoRL)}, 2026, accepted for publication.

\bibitem{xue2025reactive}
H.~Xue, J.~Ren, W.~Chen, G.~Zhang, Y.~Fang, G.~Gu, H.~Xu, and C.~Lu, ``Reactive diffusion policy: Slow-fast visual-tactile policy learning for contact-rich manipulation,'' in \emph{Proceedings of Robotics: Science and Systems}, 2025.

\bibitem{yin2023rotating}
Z.-H. Yin, B.~Huang, Y.~Qin, Q.~Chen, and X.~Wang, ``Rotating without seeing: Towards in-hand dexterity through touch,'' in \emph{Proceedings of Robotics: Science and Systems}, 2023.

\bibitem{kerr2022self}
J.~Kerr, H.~Huang, A.~Wilcox, R.~Hoque, J.~Ichnowski, R.~Calandra, and K.~Goldberg, ``Self-supervised visuo-tactile pretraining to locate and follow garment features,'' in \emph{Proceedings of Robotics: Science and Systems}, 2023.

\bibitem{heng2025vitacformer}
L.~Heng, H.~Geng, K.~Zhang, P.~Abbeel, and J.~Malik, ``{ViTacFormer}: Learning cross-modal representation for visuo-tactile dexterous manipulation,'' in \emph{Proceedings of Robotics: Science and Systems}, 2026.

\bibitem{park2026tactx}
J.~Park, S.~Bhadang, C.~Sferrazza, S.~Yi, and X.~Wang, ``Tactx: Learning shared tactile representations across diverse sensors,'' \emph{arXiv preprint arXiv:2606.31236}, 2026.

\bibitem{huang2026flexitac}
B.~Huang and Y.~Li, ``Flexitac: A low-cost, open-source, scalable tactile sensing solution for robotic systems,'' \emph{arXiv preprint arXiv:2604.28156}, 2026.

\bibitem{huang20243d}
B.~Huang, Y.~Wang, X.~Yang, Y.~Luo, and Y.~Li, ``{3D-ViTac}: Learning fine-grained manipulation with visuo-tactile sensing,'' in \emph{Proceedings of The 8th Conference on Robot Learning}, ser. Proceedings of Machine Learning Research, vol. 270.\hskip 1em plus 0.5em minus 0.4em\relax PMLR, 2025, pp. 2557--2578.

\bibitem{choi2026wild}
H.~Choi, Y.~Hou, C.~Pan, S.~Hong, A.~Patel, X.~Xu, M.~R. Cutkosky, and S.~Song, ``In-the-wild compliant manipulation with {UMI-FT},'' in \emph{2026 IEEE International Conference on Robotics and Automation (ICRA)}.\hskip 1em plus 0.5em minus 0.4em\relax IEEE, 2026.

\bibitem{oquab2023dinov2}
M.~Oquab, T.~Darcet, T.~Moutakanni, H.~Vo, M.~Szafraniec, V.~Khalidov, P.~Fernandez, D.~Haziza, F.~Massa, A.~El-Nouby \emph{et~al.}, ``{DINOv2}: Learning robust visual features without supervision,'' \emph{Transactions on Machine Learning Research}, 2024.

\bibitem{jaegle2021perceiver}
A.~Jaegle, F.~Gimeno, A.~Brock, O.~Vinyals, A.~Zisserman, and J.~Carreira, ``Perceiver: General perception with iterative attention,'' in \emph{International conference on machine learning}.\hskip 1em plus 0.5em minus 0.4em\relax PMLR, 2021, pp. 4651--4664.

\bibitem{perez2018film}
E.~Perez, F.~Strub, H.~De~Vries, V.~Dumoulin, and A.~Courville, ``Film: Visual reasoning with a general conditioning layer,'' in \emph{Proceedings of the AAAI conference on artificial intelligence}, vol.~32, no.~1, 2018.

\bibitem{lipman2022flow}
Y.~Lipman, R.~T. Chen, H.~Ben-Hamu, M.~Nickel, and M.~Le, ``Flow matching for generative modeling,'' in \emph{International Conference on Learning Representations}, 2023.

\end{thebibliography}
\end{document}